\documentclass[preprint,12pt]{elsarticle}

\usepackage{amssymb}

\usepackage[utf8]{inputenc} 
\usepackage{booktabs}       
\usepackage{amsfonts}       
\usepackage{nicefrac}       
\usepackage{microtype}      
\usepackage{amsmath}
\usepackage{amsthm}
\usepackage{amssymb}
\usepackage{algorithm}
\usepackage{algorithmic}
\usepackage{subfigure}
\usepackage{array}
\usepackage[switch]{lineno}
\usepackage{multirow}
\usepackage{color, xcolor}
\usepackage{url}
\usepackage{ulem}
\usepackage{graphicx}
\usepackage{subfigure}
\usepackage{booktabs} 
\usepackage{multirow}
\usepackage{amsmath}
\usepackage{amssymb}
\usepackage{mathtools}
\usepackage{amsthm}
\usepackage{xcolor,colortbl}
\usepackage{float}
\usepackage{amsthm}

\journal{-}

\begin{document}

\begin{frontmatter}



\author[1,2]{Xunyu Zhu}
\ead{zhuxunyu@iie.ac.cn}

\author[3]{Jian Li\corref{cor1}}
\ead{jli@bnu.edu.cn}

\author[1,2]{Can Ma}
\ead{macan@iie.ac.cn}

\author[1,2]{Weiping Wang}
\ead{wangweiping@iie.ac.cn}

\cortext[cor1]{Corresponding author}

\affiliation[1]{Institute of Information Engineering, Chinese Academy of Sciences.}

\affiliation[2]{School of Cyber Security, University of Chinese Academy of Sciences.}

\affiliation[3]{School of Artificial Intelligence, Beijing Normal University.}

\title{Improving Mathematical Reasoning Capabilities of Small Language Models via Feedback-Driven Distillation}

\begin{abstract}
Large Language Models (LLMs) demonstrate exceptional reasoning capabilities, often achieving state-of-the-art performance in various tasks. However, their substantial computational and memory demands, due to billions of parameters, hinder deployment in resource-constrained environments. A promising solution is knowledge distillation, where LLMs transfer reasoning capabilities to Small Language Models (SLMs, $\le$ 1B parameters), enabling wider deployment on low-resource devices. Existing methods primarily focus on generating high-quality reasoning rationales for distillation datasets but often neglect the critical role of data quantity and quality. To address these challenges, we propose a Feedback-Driven Distillation (FDD) framework to enhance SLMs’ mathematical reasoning capabilities. In the initialization stage, a distillation dataset is constructed by prompting LLMs to pair mathematical problems with corresponding reasoning rationales. We classify problems into easy and hard categories based on SLM performance. For easy problems, LLMs generate more complex variations, while for hard problems, new questions of similar complexity are synthesized. In addition, we propose a multi-round distillation paradigm to iteratively enrich the distillation datasets, thereby progressively improving the mathematical reasoning abilities of SLMs. Experimental results  demonstrate that our method can make SLMs achieve SOTA mathematical reasoning performance.
\end{abstract}




\end{frontmatter}



\section{Introduction}
With the rapid development of artificial intelligence, Large Language Models (LLMs) typically exhibit exceptional reasoning capabilities and consistently achieve state-of-the-art (SOTA) performance across a wide range of reasoning tasks. However, this progress has been accompanied by a dramatic increase in the size of these LLMs, leading to higher computational costs and greater memory requirements. Currently, LLMs have tens to hundreds of billions of parameters, which means they require several high-performance GPUs for deployment. These challenges pose significant obstacles to deploying LLMs in low-resource environments. 

To enable large-scale deployment, a feasible approach is to leverage knowledge distillation to transfer the reasoning capabilities of Large LLMs to Small Language Models (SLMs, $\le$ 1B), thereby enhancing the reasoning performance of SLMs. SLMs, due to their compact size, can be more widely deployed on low-resource devices. Specifically, numerous studies~\cite{hsieh-etal-2023-distilling,ho-etal-2023-large,pmlr-v202-fu23d,shridhar-etal-2023-distilling,zhu2024padprogramaideddistillationteach,ZHU2024106594} employ LLMs to construct mathematical distillation datasets. Each problem in the distillation dataset is paired with a corresponding Chain-of-Thought (CoT) or Program-of-Thought (PoT) reasoning rationale. These datasets are then used to fine-tune SLMs, significantly improving their mathematical reasoning abilities. Although these approaches can effectively enhance the mathematical reasoning abilities of SLMs, they have a potential drawback: successful distillation requires a sufficient amount of distillation data. The aforementioned methods focus primarily on generating high-quality reasoning rationales but overlook the critical impact of data quantity on mathematical distillation. On the other hand, constructing additional data can be both costly and time-consuming. Therefore, exploring how to efficiently generate a sufficient amount of high-quality data for distillation is a problem worth investigating. 

Currently, extensive research~\cite{wang-etal-2023-self-instruct,yu2024metamath,li-etal-2024-mugglemath,xu2024wizardlm} has demonstrated that LLMs can effectively generate high-quality data based on original data. However, these approaches apply a uniform generation strategy to all original data, overlooking the variations in students' performance on the original data. Jiang et al.; Lee et al.; Ying et al.~\cite{jiang-etal-2023-lion,lee-etal-2024-llm2llm,ying-etal-2024-llms} considers student performance on the original data when generating synthetic data. However, these approaches focus solely on the original data where students underperform, guiding LLMs to synthesize data based on such underperforming instances. This overlooks the potential of the original data where students perform well. To solve the problem, we propose a Feedback-Driven Distillation (FDD) framework to enhance mathematical reasoning abilities of SLMs. In the Initialization Stage, we prompt the LLM to construct a mathematical distillation dataset based on a mathematical dataset. Each question in the distillation dataset is paired with a corresponding PoT rationale. This distillation dataset is then used to fine-tune SLMs, enhancing their mathematical reasoning abilities. Next, we evaluate the SLMs' performance on the questions within the distillation dataset, categorizing them into hard questions and easy questions. Easy questions refer to those that SLMs can solve correctly, while hard questions are those they fail to solve. When dealing with easy questions, we prompt the LLMs to generate more complex questions based on these. For hard questions, the LLMs are prompted to create questions within the same domain and task type, maintaining similar levels of complexity. These newly generated questions are incorporated into the distillation dataset, expanding its scale, complexity, and diversity. Finally, we use the enriched distillation dataset to fine-tune SLMs from scratch, further improving their mathematical reasoning capabilities. Additionally, we further propose a multi-round distillation paradigm to iteratively enhance the mathematical reasoning capabilities of SLMs. In each round, we integrate the hard questions and generated questions from the previous round, leveraging the fine-tuned SLMs from that round to reclassify these questions into hard and easy categories. Subsequently, we employ LLMs to generate new questions based on the hard and easy questions, incorporating these newly generated questions into the distillation dataset. This enriched distillation dataset is then used to fine-tune SLMs from scratch, progressively improving their mathematical reasoning abilities.

In our paper, we evaluate our method on various mathematical datasets using a family of FlanT5 models. Our experimental results demonstrate that our approach can effectively enhance the mathematical reasoning capabilities of SLMs, even enabling them to achieve state-of-the-art performance in mathematical reasoning tasks. Moreover, our approach not only enhances the in-domain mathematical reasoning performance of SLMs but also significantly improves their out-of-domain mathematical reasoning capabilities.

\section{Related Work}
\subsection{Reasoning Distillation}
Reasoning Distillation refers to the process of transferring the reasoning abilities of LLMs to SLMs. A growing body of research~\cite{hsieh-etal-2023-distilling,ho-etal-2023-large,pmlr-v202-fu23d,shridhar-etal-2023-distilling,zhu2024padprogramaideddistillationteach,ZHU2024106594} has increasingly focused on this field of reasoning distillation. This approach aims to transfer the reasoning capabilities of LLMs to SLMs ($\le$ 1B), enabling more efficient deployment while maintaining strong performance. Specifically, Hsieh et al.~\cite{hsieh-etal-2023-distilling} first proposed extracting rationales from LLMs and using a multi-task learning framework to guide SLMs in learning both answers and rationales, thereby enhancing their reasoning performance. Building on this, Ho et al.~\cite{ho-etal-2023-large} discovered that fine-tuning SLMs directly with CoTs generated by LLMs could significantly improve the reasoning abilities of SLMs. Zhu et al.~\cite{zhu2024padprogramaideddistillationteach} further advanced this approach by utilizing PoTs instead of CoTs to fine-tune SLMs, achieving additional improvements in reasoning performance. This enhancement is attributed to PoTs offloading computational steps in the reasoning process to an external Python interpreter, allowing SLMs to focus solely on generating Python programs to solve problems.
Fu et al.~\cite{pmlr-v202-fu23d} introduced distribution matching distillation, which minimizes the Kullback-Leibler (KL) divergence between the output distributions of LLMs and SLMs to improve the mathematical reasoning abilities of SLMs. Similarly, Shridhar et al.~\cite{shridhar-etal-2023-distilling} trained two distilled models: a problem decomposer and a subproblem solver. When given a problem, the problem decomposer breaks it into smaller subproblems, and the subproblem solver generates CoTs to resolve these subproblems, ultimately producing the final answer. Finally, Zhu et al.~\cite{ZHU2024106594} designed a diverse reasoning format distillation framework to further enhance the mathematical reasoning capabilities of SLMs. This framework constructs a distillation dataset encompassing various reasoning formats to fine-tune SLMs effectively. Although these methods can effectively enhance the reasoning performance of SLMs, they overlook the impact of low-data regimes on the reasoning abilities of SLMs. To address this issue, we propose a feedback-driven distillation framework designed to generate additional questions, which are then used to expand the distillation dataset. This enriched dataset is subsequently leveraged to improve the reasoning performance of SLMs.

\subsection{Data Generation for LLMs}
Data Generation for LLMs refers to leveraging LLMs to generate data, thereby expanding datasets for subsequent fine-tuning tasks.  For example, Self-Instruct~\cite{wang-etal-2023-self-instruct} leverages the self-generated outputs of LLMs to bootstrap the construction of instruction-following datasets, thereby enhancing the LLMs' own ability to follow instructions. Metamath~\cite{yu2024metamath} creates a new dataset by rewriting  existing mathematical questions from multiple perspectives. This dataset is then used to fine-tune open-source LLMs, significantly improving their capacity for mathematical reasoning. MuggleMath~\cite{li-etal-2024-mugglemath} conducts an in-depth exploration of how data augmentation strategies influence the mathematical reasoning and generalization capabilities of open-source LLMs. WizardLM~\cite{xu2024wizardlm} rewrites existing instructions step by step  into more complex instructions. Furthermore, Jiang et al.; Lee et al.; Ying et al.~\cite{jiang-etal-2023-lion,lee-etal-2024-llm2llm,ying-etal-2024-llms} expand fine-tuning datasets by customizing the generation of instructions based on feedback from student LLMs. Although these methods enable student LLMs to achieve strong performance, they focus solely on instructions where student LLMs underperform, overlooking the potential of instructions where student LLMs excel. In contrast to these approaches, our method considers both types of instructions, employing distinct generation strategies for each. This dual focus enhances the complexity, scale, and diversity of the resulting instructions.

\begin{figure*}[!ht]
	\centering
	\includegraphics[width=\textwidth]{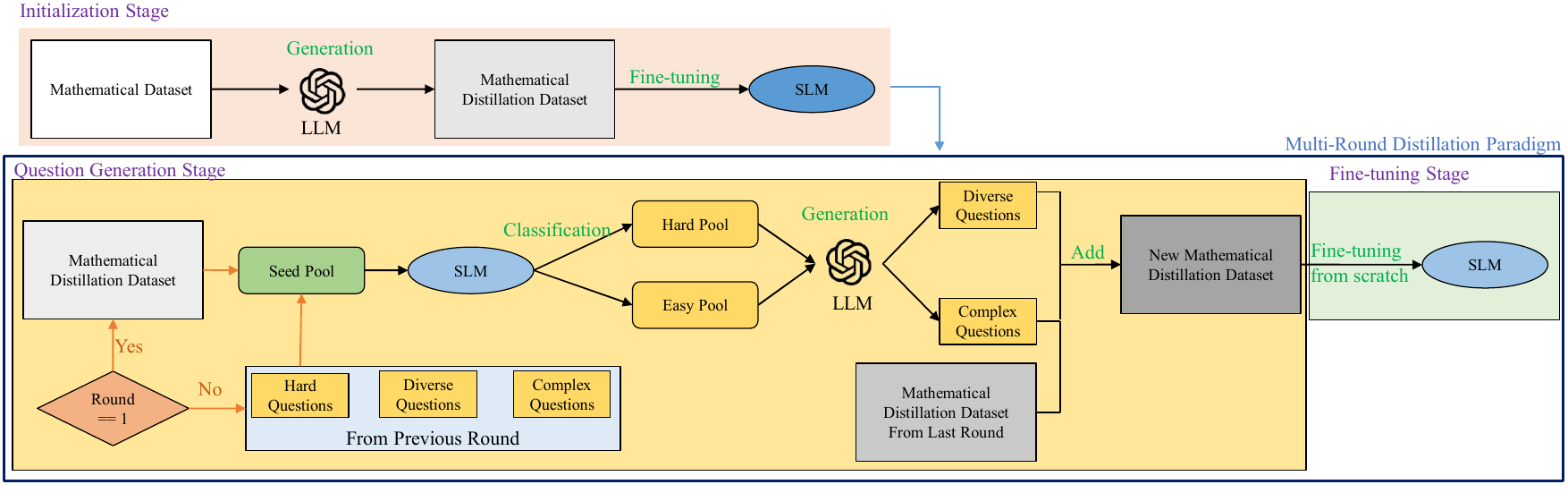}
	\caption{\textbf{The overview of our distillation method.} In the initialization stage, we prompt the LLM to create an initial distillation mathematical dataset based on the original mathematical dataset. Next, in the question generation stage, we prompt the LLM to generate more diverse questions based on hard questions and more complex questions from easy ones. These generated questions are used to expand the distillation dataset, enhancing its diversity and complexity. This expanded distillation dataset is then used to fine-tune SLMs from scratch to enhance mathematical reasoning abilities and mitigate catastrophic forgetting. Additionally, we propose a multi-round distillation paradigm to iteratively expand the dataset and further improve the mathematical reasoning capabilities of the SLMs.}
	\label{fig:fdd}
\end{figure*}

\section{Methodology}
In this work, we propose a Feedback-Driven Distillation (FDD) framework aimed at enhancing the mathematical reasoning abilities of SLMs by constructing a more complex and diverse distillation dataset. Our framework consists of three stages: (1) an initialization stage to equip SLMs with preliminary mathematical reasoning abilities, (2) a question generation stage in which LLMs generate new questions to expand and diversify the distillation dataset, and (3) a fine-tuning stage where the enriched distillation dataset is used to fine-tune SLMs, further improving their mathematical reasoning capability. Figure~\ref{fig:fdd} shows the overview of our distillation method.

\subsection{Initialization Stage}
\label{sec:init}
Firstly, we create a distillation dataset to equip SLMs with preliminary mathematical reasoning abilities. The primary goal is to enable SLMs to distinguish between hard and easy questions, providing valuable feedback to LLMs during the question generation stage. Specifically, given a mathematical dataset $\mathcal{D}$, where each entry consists of a question $q$  and a corresponding gold answer $a$, we prompt LLMs to generate a rationale for each mathematical question in the dataset. 
In this work, we guide LLMs to generate rationales in PoT format rather than CoT format. The primary advantage of PoT is that it focuses solely on generating the program needed to solve the question, while delegating the actual computational process to an external Python interpreter. This division of labor reduces the computational burden on the model. In contrast, CoT requires the model to handle the entire reasoning and computation process, which significantly increases its workload. Additionally, PoT narrows the prediction space compared to CoT, reducing the model’s expressive complexity and making the task easier for the model to handle~\cite{zhu2024padprogramaideddistillationteach}. 
To achieve our goal, we select $k$ questions from the mathematical dataset and manually construct PoTs for these questions. These questions, along with their corresponding PoTs, are then compiled into a demonstration dataset. For each question in the mathematical dataset, we combine this demonstration dataset with the question to form an instruction, which is then used to prompt the LLM to generate the corresponding PoT for that question. The PoT generation process can be represented as follows:
\begin{equation}
	p_i = \mathcal{F}_{\mathrm {M}} (q_i,\mathcal{R} ),
	\label{eq:pot_generation}
\end{equation}
where $\mathcal{R}$ is the demonstration dataset, $p$ is the PoT, $\mathrm {M}$ denotes LLMs, and $\mathcal{F}$ is the decoding function, and $i$ means the index of the question. 

After the LLM generates a corresponding PoT for each question, we use an external Python interpreter to execute the PoT, producing an answer. This answer is then compared to the gold answer from the mathematical dataset. If the two answers match, the PoT is considered correct; otherwise, it’s deemed incorrect and discarded. This process helps ensure the quality of the generated PoTs, allowing us to build a higher-quality distillation dataset, ultimately improving the mathematical reasoning performance of the fine-tuned SLMs. Finally, we get a mathematical distillation dataset $\mathcal{D}_p$, where each entry consists of a mathematical question paired with a PoT solution.

We then use this mathematical distillation dataset to fine-tune the SLMs, enabling them to acquire preliminary mathematical reasoning abilities. For each mathematical question, we combine it with the prompt $pr$, "Let's generate a python program to solve the question." to create an instruction as the input, with the corresponding PoT as the output. The fine-tuning loss function can be represented as follows:
\begin{equation}
	\mathcal{L} = - \sum_{i=1}^{N} \sum_{t=1}^{T} \log P({p}^i_t \mid {p}^i_{< t}, q^i, pr),
	\label{eq:fine_tuning}
\end{equation}
where $N$ represents the quantity of data within the mathematical distillation dataset, and $p_{:T}$ denotes the sequence of PoTs. When faced with a new mathematical question, the fine-tuned SLMs generate a PoT and then use a Python interpreter to execute it, producing the corresponding answer.

\subsection{Question Generation Stage}
Previous studies~\cite{xu2024wizardlm,li-etal-2024-mugglemath,yu2024metamath} have shown that the complexity and diversity of fine-tuning datasets play a significant role in shaping the reasoning performance of SLMs. Building on this insight, our work seeks to enhance the complexity and diversity of the mathematical distillation dataset to improve the mathematical reasoning capabilities of SLMs. 

Specifically, we use the fine-tuned SLM from the Initialization Stage to perform inference on each data in the mathematical distillation dataset, generating corresponding answer. We then compare the generated answer with the gold answer. If the generated answer matches the gold answer, we consider the question to be easy, meaning the SLM have successfully learned how to solve it. Conversely, if the generated answer does not match, we classify the question as hard, indicating that the SLM have yet to master it. In the teaching process, when students have learned how to solve a particular question, teachers often build a more challenging question based on the original one. This approach helps students deepen their understanding of the relevant knowledge. Conversely, if a student struggles with a question, the teacher typically generates additional questions of similar knowledge scope and difficulty, allowing the student to learn from repeated exposure to related questions. Inspired by this approach, we introduce a feedback-driven question generation strategy. For an easy question, we prompt the LLM to generate a new question by increasing the complexity of the original. In contrast, for a hard question, we use a method similar to the self-instruct approach~\cite{wang-etal-2023-self-instruct}, prompting the LLM to treat the hard question as a seed for generating a similar question. The question generation process can be represented as follows:
\begin{equation}
	q_{new} = \mathcal{F}_{\mathrm {M}} (q_{old}, \mathcal{I} ),
\end{equation}
where $q_{new}$ is the new generated question, $q_{old}$ is the original question, and $\mathcal{I}$ is the instruction that prompts the LLM to generate a new question, and it is provided in Appendix~\ref{sec:instructforqg}.   

After getting a new question, we follow the same approach as outlined in Section~\ref{sec:init} to prompt the LLM to generate a PoT for the question. We then use a Python interpreter to extract the corresponding answer from this PoT. To ensure the quality of both the PoT and the answer, we prompt the model to generate multiple PoTs. Then, we extract the corresponding answers from these PoTs and apply a voting mechanism to select the answer with the highest number of votes as the gold answer. The PoTs that correspond to the gold answer are considered correct and are added, along with the question, to the distillation dataset. This approach helps to enhance the complexity and diversity of the distillation dataset.

\subsection{Fine-tuning Stage}
After expanding the scale of the distillation dataset, we use this dataset to fine-tune the SLMs to improve their mathematical reasoning performance. The fine-tuning process of the SLMs follows the method outlined in Section~\ref{sec:init}. For each question in the distillation dataset, we combine it with the prompt "Let's generate a Python program to solve the question." to construct an input instruction.  This input instruction is then used to fine-tune the SLMs, enabling them to generate the corresponding PoT for each question. The fine-tuning loss function can also be represented by Equation~\ref{eq:fine_tuning}. Notably, we use the distillation dataset to fine-tune the SLMs from scratch to avoid catastrophic forgetting.

\subsection{Multi-Round Distillation Paradigm}
In the subsection, we further propose a multi-round distillation paradigm to iteratively improve the mathematical reasoning abilities of SLMs. The multi-round distillation paradigm enables the LLM to stay up-to-date with the learning state of the SLM, allowing it to generate questions that are better aligned with the SLM's current progress.

Specifically, we first apply methods from the Initiation stage to construct an initial distillation dataset. This dataset is then used to fine-tune SLMs, enabling them to acquire preliminary mathematical reasoning abilities. In turn, this equips the SLMs to provide valuable feedback to the LLM, guiding it in generating customized questions. After initializing the SLMs, we add all data from the initial distillation dataset to the seed pool in the next round. Using the initialized SLM, we then categorize the data in the seed pool as either hard or easy questions, placing them accordingly into a hard pool and an easy pool. For data in the hard pool, the LLM generates questions of similar type and difficulty. For data in the easy pool, the LLM generates more challenging questions. These newly generated questions are added to the distillation dataset, which is then used to fine-tune the SLMs from scratch. In each subsequent round, we exclude the easy questions from the previous round from the seed pool. Instead, we add only the newly generated questions and the hard questions to the seed pool. This approach avoids redundancy, as the model has already learned the relevant knowledge from the easy questions in the previous round, and the more complex questions derived from them are already included in the seed pool. This strategy also reduces dataset size, lowers data generation costs, and improves fine-tuning efficiency. Following the same method, the LLM generates new questions, which are added to the distillation dataset, allowing us to fine-tune the SLMs from scratch.

\section{Experiments}

\subsection{Dataset}
In our work, we use the GSM8K training set~\cite{abs-2110-14168}, which contains 7,473 examples, as the mathematical reasoning dataset for our experiments. We prompt the LLM to construct a mathematical distillation dataset based on this GSM8K training set. Subsequently, we fine-tune the SLMs on this mathematical distillation dataset to enhance their mathematical reasoning capabilities. Furthermore, we evaluate the mathematical reasoning performance of the SLMs on the GSM8K test set, which contains 1,319 examples. Additionally, to assess the transferability of the SLMs' mathematical reasoning abilities, we evaluate them on several well-known mathematical reasoning datasets, including ASDiv~\cite{miao-etal-2020-diverse} with 2,300 samples, SVAMP~\cite{patel-etal-2021-nlp} with 1,000 samples, and MultiArith~\cite{roy-roth-2015-solving} with 180 samples.

\subsection{Implementation Details}
In our work, we employ ChatGPT (gpt-3.5-turbo) as the LLM to construct a mathematical distillation dataset. For our small language models (SLMs), we use FlanT5~\cite{flant5} models with parameter counts ranging from 60M to 760M.  In the initialization stage, we prompt the LLM to generate 4 PoTs for each question in the GSM8K training dataset, creating the foundation for the mathematical distillation dataset, which we then use to fine-tune the SLMs. We fine-tune the SLMs over 10 epochs, with a learning rate of 5e-4 and a batch size of 32. To further enhance the mathematical reasoning capabilities of the SLMs, we employ a three-round multi-round distillation paradigm. In each round, we also generate 4 PoTs for each new question, and the fine-tuning settings for the SLMs remain consistent with those used in the initialization stage. Additionally, during the generation of math questions, we set the LLM’s temperature to 1, while for PoT generation, we set it to 0.7.

\begin{table*}[!ht]
	\centering
	\scriptsize
	\begin{tabular}{c|c|cccc|c}
		\toprule 
		\textbf{Models} & \textbf{\#Params} & \textbf{GSM8K} & \textbf{ASDiv} & \textbf{SVAMP} & \textbf{MultiArith} & \textbf{AVG}\\
		\midrule 
		\rowcolor[rgb]{0.93,0.93,0.93}
		\multicolumn{7}{l}{\textit{Proprietary Large Language Models}} \\
		GPT-4~\cite{openai2024gpt4technicalreport} & - & 92.0 & 91.3 & 93.1 & - & 92.13\\
		ChatGPT & -  & 80.8 & 87.3 & 83.0 & - & 83.7\\
		Claude-2 & - & 85.2 & - & - & - & 85.2\\
		PaLM-2~\cite{anil2023palm2technicalreport} & 540B & 80.7 & - & - & - & 80.7\\
		\midrule 
		\rowcolor[rgb]{0.93,0.93,0.93}
		\multicolumn{7}{l}{\textit{Open-Source Large Language Models}} \\
		Llama-2~\cite{touvron2023llama2openfoundation} & 7B & 13.3 & 50.7 & 38.0 & - & 34\\
		CodeLLaMA~\cite{rozière2024codellamaopenfoundation} & 7B & 34.0 & 61.4 & 59.0 & - & 51.46\\
		Platypus-2~\cite{lee2024platypusquickcheappowerful} & 7B & 14.4 & 47.9 & 36.7 & - & 33\\        WizardMath~\cite{luo2023wizardmathempoweringmathematicalreasoning} & 7B & 54.9 & 59.1 & 57.3 & - & 57.1\\
		TORA~\cite{gou2024tora} & 7B & 68.8 & 73.9 & 68.2 & - & 70.3 \\
		\midrule 
		\rowcolor[rgb]{0.93,0.93,0.93}
		\multicolumn{7}{l}{\textit{Fine-tuned Small Language Models}} \\
		Ho et al.~\cite{ho-etal-2023-large} & 0.3B & 3.11 & - & - & - & 3.11\\
		Fu et al.~\cite{pmlr-v202-fu23d} & 0.76B & 20.2 & 23.8 & 20.4 & 38.5 & 25.72\\
		Shridhar et al.~\cite{shridhar-etal-2023-distilling} & 0.77B & 17.89 & - & 18.14 & - & 18.01\\
		Zhu et al.~\cite{zhu2024padprogramaideddistillationteach} & 0.77B & 39.2 & 51.2 & 48.2 & 79.2 & 54.45\\
		Zhu et al.~\cite{ZHU2024106594} & 0.77B & 42.45 & 52.81 & 49.59 & 85.5 & 57.58\\
		\midrule 
		\rowcolor[rgb]{0.93,0.93,0.93}
		\multicolumn{7}{l}{\textit{Our fine-tuned Small Language Models}} \\
		FlanT5-Small & 0.06B & 2.1 & 2.8 & 2.1 & 4.0 & 2.75\\
		(+) FDD & & \textbf{29.87} & \textbf{52.86} & \textbf{43.4} & \textbf{77.16} & \textbf{50.82}\\
		\hline
		FlanT5-Base & 0.25B & 3.0 & 4.2 & 3.8 & 7.0 & 4.5\\
		(+) FDD & & \textbf{40.25} &\textbf{58.44}& \textbf{54.3} & \textbf{87.83} & \textbf{60.20}\\
		\hline
		FlanT5-Large & 0.76B & 6.9 & 10.1 & 6.8 & 13.0 & 9.2\\
		(+) FDD & & \textbf{49.43} & \textbf{64.88} & \textbf{61.9} & \textbf{94.0} & \textbf{67.55}\\
		\bottomrule 
	\end{tabular}
	\caption{\textbf{Overall Test Set Performance.} }
	\label{tab:main_results}
\end{table*}

\subsection{Main Results}
Table~\ref{tab:main_results} shows the main experimental results. Based on the results, we observe that: (1) Our method enables SLMs to achieve State-of-the-Art (SOTA) mathematical reasoning performance. In Table~\ref{tab:main_results}, we demonstrate how our method enhances the mathematical reasoning capabilities of the FlanT5 models—Small (60M), Base (250M), and Large (760M). We evaluate these SLMs on the in-domain GSM8k test set, where our approach (FDD) improves FlanT5-small to an accuracy of 29.87\%, FlanT5-Base to 40.25\%, and FlanT5-Large to 49.43\%. Compared to other distillation methods focused on SLMs ($\le$1B parameters), our approach achieves SOTA performance in mathematical reasoning, even surpassing some open-source LLMs. These experimental results validate the effectiveness of our method. (2) Our method also demonstrates good transferability. Specifically, we further evaluate the FlanT5 models, fine-tuned with our method, on out-of-domain mathematical reasoning test sets, including ASDiv, SVAMP, and MultiArith. Table~\ref{tab:main_results} shows that FDD enables FlanT5-small to reach 52.86\% accuracy on ASDiv, 43.4\% on SVAMP, and 77.16\% on MultiArith; FlanT5-Base to achieve 58.44\%, 54.3\%, and 87.83\%, respectively; and FlanT5-Large to attain 64.88\%, 61.9\%, and 94.0\% on these benchmarks. Our method outperforms other distillation techniques on these out-of-domain datasets, further underscoring its strong transferability. (3) Model size of SLMs impacts the mathematical reasoning performance achieved by our method. As illustrated in Table~\ref{tab:main_results}, FlanT5-small with 0.06B parameters achieves an average accuracy of 50.85\% across these mathematical reasoning datasets, FlanT5-Base with 0.25B parameters reaches 60.20\%, and FlanT5-Large with 0.76B parameters achieves 67.55\%. These results indicate that model size significantly influences mathematical reasoning performance in our method, with larger models consistently achieving better results. In summary, these findings confirm the effectiveness, transferability, and scalability of our method in enhancing mathematical reasoning performance of
SLMs.

\begin{table*}[!ht]
	\centering
	\scriptsize
	\begin{tabular}{@{}l|c|ccccc@{}}
		\toprule
		\textbf{Strategy}    & \textbf{Data Size} & \textbf{GSM8K} & \textbf{ASDiv} & \textbf{SVAMP} & \textbf{MultiArith} & \textbf{AVG}   \\ \midrule
		None & 0  & 3.0   & 4.2   & 3.8   & 7.0        & 4.5   \\ \midrule
		(+) Initialization Dataset & 4,620     & 20.09 & 42.84 & 36.7  & 44.83      & 36.11 \\ \midrule
		(+) Complex Questions  & 1,666     & 23.73 & 45.80 & 42.0  & 51.0       & 40.63 \\ 
		(+) Diverse Questions  & 1,928     & 24.33 & 47.32 & 39.9  & 56.66      & 42.05 \\ \midrule
		(+) All Questions      & 3,594     & 28.05 & 48.47 & 42.8  & 56.66      & 43.99 \\ \bottomrule
	\end{tabular}
	\caption{\textbf{Effect of Question Generation Strategies.} The experimental results show that the combination of the complex and diverse question generation strategies can improve the mathematical reasoning performance of SLMs.}
	\label{tab:effect_strategy}
\end{table*}

\subsection{Effect of Question Generation Strategies}
In this section, we explore the effect of different question generation strategies on the mathematical reasoning performance of SLMs. To achieve this, we first prompt the LLM to construct an initial mathematical distillation dataset. This initialization dataset is then used to fine-tune FlanT5-base from scratch. Next, we use the fine-tuned FlanT5-base model to classify the questions in the distillation dataset into easy and hard questions. For easy questions, we prompt the LLM to generate more complex questions based on them. For hard questions, we prompt the LLM to generate diverse questions based on the originals. We then employ three strategies for constructing the distillation dataset: (1) adding the complex questions to the distillation dataset, (2) adding the diverse questions to the distillation dataset, and (3) adding all questions—including both easy and hard questions—into the dataset. The extended distillation dataset is then used to fine-tune FlanT5-base from scratch again. By evaluating the mathematical reasoning performance of the resulting SLMs, we analyze the effect of different question generation strategies on performance. Additionally, to simplify the analysis, we ensure that each question in the distillation dataset contains only a single PoT in this study.

Table~\ref{tab:effect_strategy} shows the experimental results about the study. Based on these results, we find that: (1) The initialization stage is crucial as it significantly improves the mathematical reasoning performance of the SLM. This is because the questions in the initialization distillation dataset are manually created, ensuring high quality and diversity. Additionally, we can obtain the gold answers for these questions, and by comparing them with the answers extracted from PoT, we can further verify the quality of the generated PoT. Overall, the experimental results demonstrate that the initialization stage plays a vital role in enhancing the mathematical reasoning abilities of the SLM. (2) Both the complex and diverse question generation strategies can improve the SLM's mathematical reasoning abilities. Specifically, the complex question generation strategy achieved 23.73\% accuracy on the GSM8k test set and 40.63\% average accuracy across all test sets. The diverse question generation strategy performed slightly better, achieving 24.33\% accuracy on GSM8K and 42.05\% average accuracy across all test sets. In other words, compared to the performance of the SLM in the initialization stage, both strategies lead to improved mathematical reasoning capabilities. (3) The combination of the complex and diverse question generation strategies further boosts the SLM's mathematical reasoning performance. Table~\ref{tab:effect_strategy} indicates that when all the newly generated questions are added to the distillation dataset, training the expanded dataset on FlanT5-base results in a 28.05\% accuracy on GSM8K and 43.99\% average accuracy across all test sets, which significantly surpasses the performance achieved by either strategy alone. We attribute this improvement to the higher complexity, better diversity, and larger scale of the dataset extended by combining the two strategies. Thus, in our work, we combine both strategies to generate new questions, expanding the distillation dataset.

\begin{figure}[!ht]
	\includegraphics[width=\columnwidth]{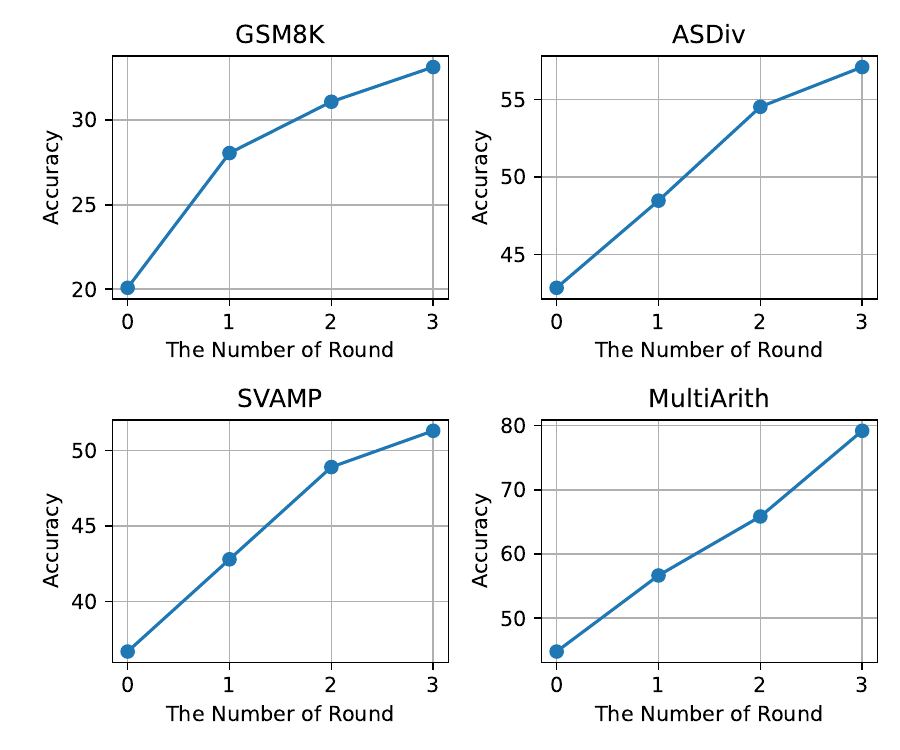}
	\caption{\textbf{Effect of the Number of Rounds.} The experimental results demonstrate the effectiveness of the multi-round distillation paradigm, while additional rounds further enhance the mathematical reasoning performance of SLMs.}
	\label{fig:effect_round}
\end{figure}

\subsection{Effect of the Number of Rounds}
In this subsection, our primary objective is to investigate the impact of the number of rounds on the mathematical reasoning performance of SLMs. Specifically, we begin by fine-tuning the SLMs during the initialization stage to equip them with basic mathematical reasoning abilities. In the experiment, the initialization stage is referred to as "round 0." Next, we employ a multi-round distillation paradigm to further enhance the mathematical reasoning performance through up to three rounds. We record the SLM's performance in mathematical reasoning at each round. Based on these performance, we explore the effect of the number of rounds on the SLM's capabilities. For the sake of simplicity in our analysis, we use FlanT5-base as the SLM in this experiment and ensure that each question in the distillation dataset has only one PoT.

Figure~\ref{fig:effect_round} presents the experimental results of this study. Based on these results, we find that: (1) More rounds improve the in-domain mathematical reasoning performance of SLMs. Specifically, Figure~\ref{fig:effect_round} shows that as the rounds progress from round 0 to round 3, FlanT5-base’s mathematical reasoning performance on the GSM8k test set gradually increases. We attribute this outcome to the fact that with each additional round, more generated questions are added to the distillation dataset, thereby increasing its size, diversity, and complexity. (2) More rounds also enhance the out-of-domain mathematical reasoning performance of SLMs. Specifically, when we evaluate SLMs using out-of-domain datasets such as ASDiv, SVAMP, and MultiArith, we observe that as the number of rounds increases, the mathematical reasoning performance improves. These findings validate the effectiveness of our multi-round distillation paradigm, demonstrating that additional rounds lead to better overall mathematical reasoning performance for SLMs.

\begin{figure}[!ht]
	\includegraphics[width=\columnwidth]{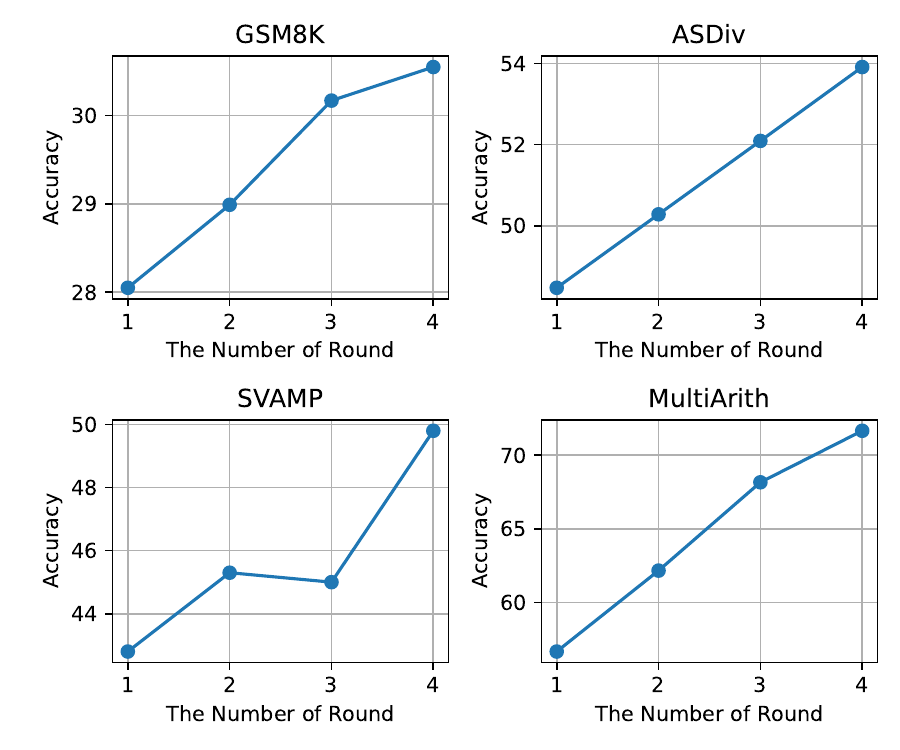}
	\caption{\textbf{Effect of the Number of Reasoning Paths.} The experimental results show that more reasoning paths can make SLMs have better mathematical reasoning performance.}
	\label{fig:effect_reason_num}
\end{figure}

\subsection{Effect of the Number of Reasoning Paths}
In this subsection, our primary objective is to investigate the impact of the number of reasoning paths generated for each question during the question generation stage on the mathematical reasoning performance of the SLM. To achieve this goal, we design the following experiment. For ease of analysis, we begin by ensuring that each question in the distillation dataset during the initialization stage contains only one reasoning path. We then use this distillation dataset to fine-tune the flanT5-base model. Afterward, we generate math questions based on the fine-tuned flanT5-base. For each generated question, we create a set of reasoning paths, ranging from 1 to 4 paths, and add both the questions and their corresponding reasoning paths to the distillation dataset. Subsequently, we fine-tune the flanT5-base model from scratch using the expanded distillation dataset. Finally, we evaluate the mathematical reasoning performance of the fine-tuned flanT5-base on the GSM8K, ASDiv, SVAMP, and MultiArith datasets. Through these evaluations, we aim to investigate how the number of reasoning paths affects the mathematical reasoning performance of the SLMs.

Figure~\ref{fig:effect_reason_num} presents the experimental results of this study. Based on the findings, we observe that (1) increasing the number of reasoning paths improves the in-domain mathematical reasoning performance of SLMs. Specifically, Figure~\ref{fig:effect_reason_num} demonstrates that as the number of reasoning paths increases, the performance of FlanT5-base on GSM8K continues to improve. We attribute this result to the enhanced diversity of the distillation dataset, which allows SLMs to learn more effectively. (2) A greater number of reasoning paths also enhances SLM's out-of-domain mathematical reasoning performance. Specifically, as the number of reasoning paths increases, FlanT5-base achieves better results on out-of-domain datasets such as ASDiv, SVAMP, and MultiArith. These findings support the rationale that generating more reasoning paths improves the mathematical reasoning performance of SLMs.

\begin{figure}[!ht]
	\centering
	\includegraphics[width=\columnwidth]{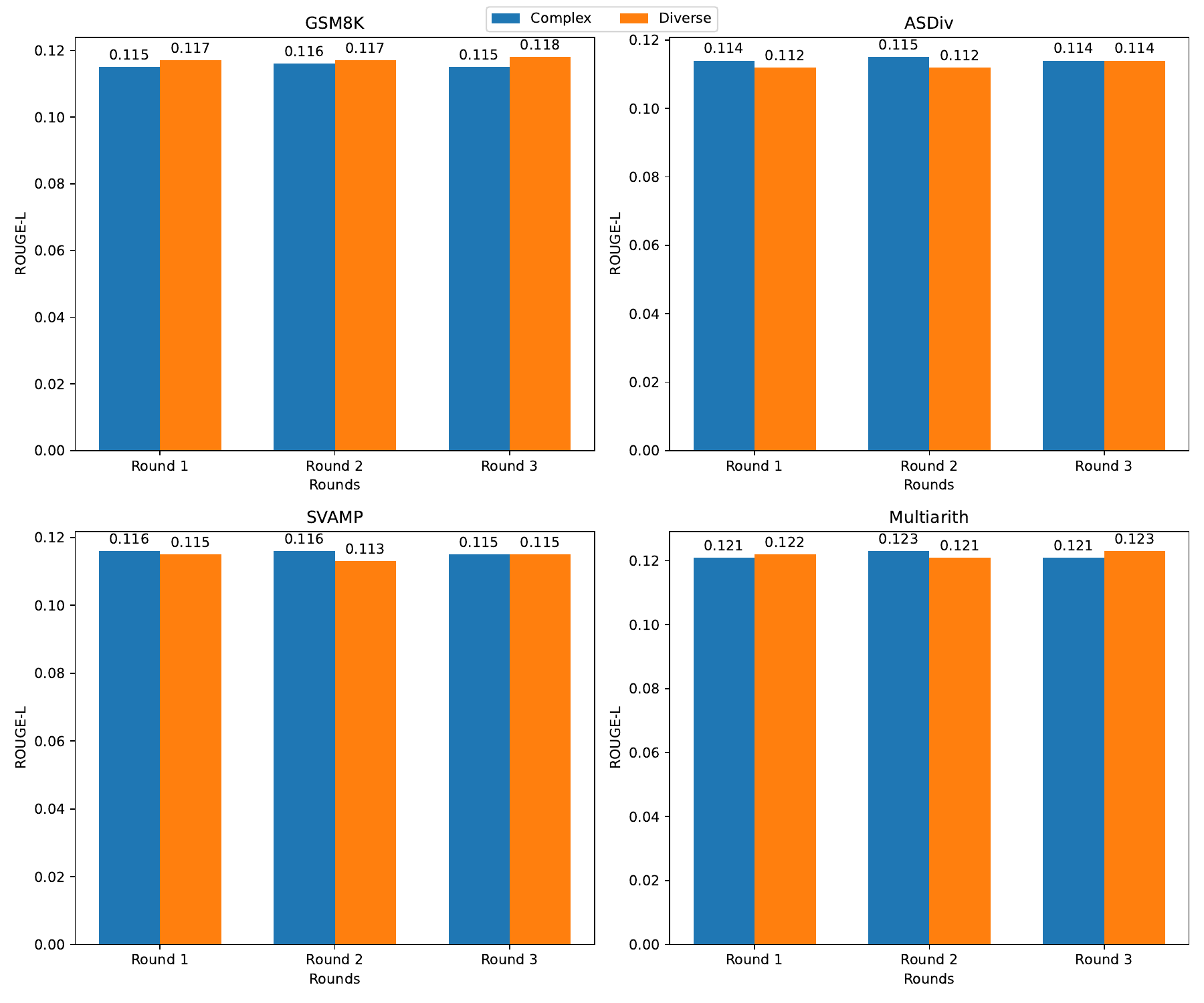}
	\caption{\textbf{Analysis for Data Leakage.} The experimental results demonstrate that in our approach, the data generated by the LLM exhibits minimal correlation with the data in the test datasets, effectively eliminating the impact of data leakage on the mathematical reasoning performance of the SLM.}
	\label{fig:data-leakage}
\end{figure}
\subsection{Analysis for Data Leakage}
When using LLMs to generate data, a potential issue that may arise is data leakage. Data leakage refers to the possibility that test data was inadvertently used during the training of the LLM, which may result in the model generating questions similar to those in the test set. In this subsection, we investigate whether our approach suffers from data leakage. To achieve this, we design a simple experiment. During the initialization stage, we prompt the LLM to construct an Initialization Mathematical Distillation Dataset based on the GSM8K training dataset, where each question is paired with a corresponding  PoT. We then fine-tune the FlanT5-base model using this distilled dataset. Following this, we implement a multi-round distillation paradigm for three rounds. At each round, we evaluate the similarity between the datasets generated in that round and the test dataset to determine the extent of data leakage. In this subsection, we compute the ROUGE-L similarity between all pairs of questions from the two datasets and take the average value as the overall similarity between the datasets. This similarity metric can be formalized as:
\begin{equation}
	\operatorname{Sim}(D_\mathcal{G}, D_\mathcal{T})=\frac{1}{m \times n} \sum_{i=1}^m \sum_{j=1}^n \operatorname{ROUGE}-\mathrm{L}\left(d_\mathcal{G}^i, d_\mathcal{T}^j\right),
\end{equation}
where $D_\mathcal{G}$ and $D_\mathcal{T}$ represent the datasets composed of generated questions and the test dataset, respectively, while $m$ and $n$ denote the sizes of the generated dataset and the test dataset, respectively. Thus, the time complexity of the similarity metric is $O\left(m \cdot n \cdot L_1 \cdot L_2\right)$, where $L_1$ and $L_2$ represent the lengths of the two questions involved in each ROUGE-L computation. To save computation time, we sample only 30\% of the generated dataset to calculate its similarity to the test datasets.

Figure~\ref{fig:data-leakage} shows the experimental results of the analysis. Based on the experiments, we observe that across these three rounds, the similarity between datasets generated from hard questions and those generated from easy questions, compared to the test datasets—including the GSM8K test set, ASDiv, SVAMP, and MultiArith—falls within the range of 0.11 to 0.12. This implies that the data generated by the LLM in our approach is essentially uncorrelated with the test datasets. This further demonstrates that our method improves the mathematical reasoning performance of SLMs by enhancing their mathematical reasoning abilities, effectively eliminating the interference of data leakage in our research.

\section{Conclusion}
In this work, we introduce the Feedback-Driven Distillation (FDD) framework to enhance the mathematical reasoning capabilities of SLMs. By leveraging an iterative process involving the generation of complex and diverse mathematical questions, we successfully expand and enrich distillation datasets. This allow for the progressive fine-tuning of SLMs, enabling them to achieve state-of-the-art performance on both in-domain and out-of-domain mathematical reasoning tasks. Furthermore, our analysis verify that these improvements were achieved without data leakage, underscoring the effectively of our method. However, our method still has some limitations. For instance, it relies on LLMs to generate questions, which increases the cost of using LLMs. Additionally, in each round, the SLM requires fine-tuning from scratch, which further adds to the training cost. In the future, we want to design more efficient mathematical reasoning methods to address these challenges and further enhance the mathematical reasoning capabilities of SLMs.

\section*{Acknowledgments}
The work of Jian Li is supported partially by National Natural Science Foundation of China (No. 62106257). 

\bibliographystyle{elsarticle-num}
\bibliography{ref}{}

\newpage
\appendix

\section{Instructions for Question Generation}
\label{sec:instructforqg}

\begin{table}[!ht]
	\centering
	\small
	\begin{tabular}{p{\linewidth}}
		\toprule
		I want you act as a Math Question Creator. \\ Your goal is to draw inspiration from the Given Math Question to create a more challenging math question by increasing the complexity of the Given Math Question. \\ The created math question should belong to the same domain and the same task type as the Given Math Question. \\ The Created Math Question must be reasonable and can be understood and solved by humans. \\
		Given Math Question: \{\textit{easy question}\} \\
		Created Math Question:
		\\
		\bottomrule
	\end{tabular}
	\caption{Instruction for creating new math question based on the easy question.}
\end{table}

\begin{table}[!ht]
	\centering
	\small
	\begin{tabular}{p{\linewidth}}
		\toprule
		I want you act as a Math Question Creator. \\ Your goal is to draw inspiration from the Given Math Question to create a new math question. \\ The created math question should belong to the same domain and the same task type as the Given Math Question. \\ The difficulty level of the Created Math Question should be similar to that of the Given Math Question. The Created Math Question must be reasonable and can be understood and solved by humans. \\
		Given Math Question: \{\textit{hard question}\} \\
		Created Math Question:
		\\
		\bottomrule
	\end{tabular}
	\caption{Instruction for creating new math question based on the hard question.}
\end{table}

\end{document}